\documentclass{article}

\DeclareSymbolFont{bbold}{U}{bbold}{m}{n}
\DeclareSymbolFontAlphabet{\mathbbold}{bbold}
\usepackage{enumerate}
\usepackage{amsmath}
\usepackage{amsfonts}
\usepackage{amssymb}
\usepackage{amsbsy}
\usepackage{bbm}
\usepackage{amsthm}
\usepackage{dsfont}
\usepackage{algorithm}
\usepackage{algpseudocode}
\usepackage{mathrsfs}
\usepackage{paralist}
\usepackage{epsfig}
\usepackage{subfig}
\usepackage{makeidx}

\newtheorem{remark}{Remark}
\newtheorem{corollary}{Corollary}
\newtheorem{lemma}{Lemma}
\newtheorem{theorem}{Theorem}

\newtheorem{definition}{Definition}
\newtheorem{fact}{Fact}

\newcommand \E {\mathop{\mbox{\ensuremath{\mathbb{E}}}}\nolimits}

\renewcommand \Pr {\mathop{\mbox{\ensuremath{\mathbb{P}}}}\nolimits}

\newcommand{\tuple}[1]{\left\langle #1\right\rangle }

\newcommand{\cset}[2]{\left\{\, #1 ~\middle|~ #2 \,\right\} }

\newcommand\Simplex {{\mathbbold{\Delta}}}
\newcommand\Reals {{\mathds{R}}}

\newcommand \CA {{\mathcal{A}}}

\newcommand \CH {{\mathcal{H}}}

\newcommand \CM {{\mathcal{M}}}

\newcommand \CS {{\mathcal{S}}}

\newcommand \bu {{\mathbf{u}}}

\newcommand \defn {\mathrel{\triangleq}}

\DeclareMathAlphabet{\mathpzc}{OT1}{pzc}{m}{it}

\newcommand \hstat {\phi}
\newcommand \stat {\Phi}

\newcommand \bel {\xi}
\newcommand \Bels {\Xi}
\newcommand \pol {\pi}
\newcommand \Pols {\Pi}
\newcommand \PolsHD {\Pols^{\mathrm{D}}}
\newcommand \PolsMD {\Pols_1^{\mathrm{D}}}
\newcommand \PolsHS {\Pols^{\mathrm{S}}}
\newcommand \PolsMS {\Pols_1^{\mathrm{S}}}

\newcommand \mdp {\mu}
\newcommand \MDPs {\CM}

\newcommand \util {U}
\newcommand \init {\beta}

\newcommand \Eb {\E_\bel}

\newcommand \Ebp {\E_\bel^\pol}

\newcommand \bv {\mathbf{v}}

\newcommand \rew {\rho}
\newcommand \Qk {q_k}

\newcommand\ind[1]{\mathop{\mbox{\ensuremath{\mathbb{I}}}}\left\{#1\right\}}

\newcommand\dd{\,\mathrm{d}}

 \newcommand{\mc}[1]{\ensuremath\mathcal{#1}}

 \newcommand{\bi}{\begin{itemize}}
 \newcommand{\ei}{\end{itemize}}

\def\clap#1{\hbox to 0pt{\hss#1\hss}}

\usepackage{nips14submit_e}

\usepackage{natbib}
\title{Generalised Entropy MDPs and Minimax Regret}
\author{Emmanouil G. Androulakis \\
G{\"o}teborgs Universitet \\
Gothenburg, Sweden\\
\texttt{emmanouil.an@outlook.com} 
\and
\bf Christos Dimitrakakis\\
Chalmers University of Technology\\
Gothenburg, Sweden\\
\texttt{chrdimi@chalmers.se }
}

\nipsfinalcopy

\begin{document} 
\maketitle

\begin{abstract} 
Bayesian methods suffer from the problem of how to specify prior beliefs. One interesting idea is to consider worst-case priors. This requires solving a stochastic zero-sum game. In this paper, we extend well-known results from bandit theory in order to discover minimax-Bayes policies and discuss when they are practical.
\end{abstract} 

\section{Introduction}
\label{sec:introduction}
In this work, we consider the problem of a Bayesian agent interacting with a Markov decision process (MDP). However the agent is unsure of how to select its prior distribution and so it prefers a choice that is safe against a potentially adversarial Nature. The problem is how to select such a policy in a computationally efficient manner.  We first recall the definition of an MDP.
\begin{definition}
  A Markov decision process $\mdp \in \MDPs$ on a state-action space $\CS \times \CA$ is a tuple $\tuple{\CS, \CA, P, \rew, T}$ where $\CS$ is a set of states, $\CA$ is a set of actions, $P$ is a transition kernel, such that
  \[
  s_{t+1} \mid s_t = s,\ a_t = a \sim P^{s,a},
  \]
  where $r_t = \rew(s_t)$ and $T$ is a (potentially random) horizon.
\end{definition}
The agent's utility is an additive function of individual rewards $r_t \in \Reals$
\begin{equation}
  \label{eq:utility-function}
  U \defn \sum_{t=1}^T r_t.
\end{equation}
For simplicity, we can assume that the reward function is known to the agent and then for a finite state space $\rho$ can be taken to be a fixed vector. For any MDP $\mdp \in \MDPs$ and policy $\pol \in \Pols$, the expected utility is
$\E^\pol_\mdp (U)$, while the conditional expected utility is called the \emph{value function}:
\begin{equation}
  \label{eq:value-function}
  V^\pol_\mdp(s) \defn \E_\mdp^\pol \left(\sum_{t=1}^T r_t ~\middle|~ s_t = s \right).
\end{equation}
For finite MDPs, and a $(1 - \gamma)$-geometrically distributed horizon $T$, the value function can be written as a vector $V_\mdp^\pol = (I - \gamma P^\pol_\mdp)^{-1} \rho$, where $P^\pol_\mdp$ is the Markov chain induced on the MDP by the policy $\pol$.
Since we are uncertain about $\mdp$, we can instead define a prior distribution $\bel$ on $\MDPs$. 
Then 
\begin{equation}
  \label{eq:belief-value-function}
  V^\pol_\bel \defn \int_\MDPs V^\pol_\mdp \dd{\bel}(\mdp),
\end{equation}
will denote the value function under the particular distribution $\bel$ on the MDPs.
Finally, for a given probability measure $\init$ on $\CS$,
which can be taken to represent a starting state distribution,
we define the utility of a particular policy $\pol$ to be:
\begin{equation}
  \label{eq:utility}
  \util(\bel, \pol) \defn \Ebp( \init^\top V^\pol_\mdp)
  = \int_\MDPs \init^\top V^\pol_\mdp \dd{\bel}(\mdp).
\end{equation}

There are two possible ways to interpret the measure $\bel$, depending on how it is chosen. If $\bel$ is selected by the agent selecting $\pol$, then it corresponds to the subjective belief of the decision maker about which is the most likely MDP \emph{a priori}. Then $U(\bel, \pol)$ corresponds to the expected utility of a particular policy under this belief. 
Let
\begin{equation}
  U^*(\bel) \defn \sup_{\pol \in \Pols} U(\bel, \pol)
  \label{eq:bayes-optimal-value}
\end{equation}
denote the Bayes-optimal utility for a belief. 
We recall the fact that this is a convex function~\citep[c.f.][]{Degroot:OptimalStatisticalDecisions}. By definition, and due to convexity, the following bounds hold:
\begin{equation}
  \label{eq:convex-value-function}
  \util(\bel, \pol) \leq U^*(\bel) \leq \int_\bel U^*(\mdp) \dd{\bel}(\mdp), \qquad \forall \pol \in \Pols.
\end{equation}
In the above, the left hand side is the utility of an arbitrary policy, while the right side can be seen as the expected utility we would obtain if the true MDP was revealed to us.

The second view of $\bel$ is to assume that the MDP is \emph{actually} drawn randomly from the distribution $\bel$. If this is known, then the subjective value of a policy is equal to its true expected value. However, it is more interesting to consider the case where Nature selects some $\bel$ in an arbitrary way from a \emph{set} of possible priors $\Bels$. Then we wish to find a policy $\pol^*$ achieving:
\begin{equation}
  \label{eq:maximin-policy}
  \max_{\pol \in \Pols} \min_{\bel \in \Bels} U(\bel, \pol).
\end{equation}
One basic open question is whether the maximum exists. This is answered in the affirmative when the game between nature and the agent has a value, i.e.
\begin{equation}
  \label{eq:game-value}
  U^* = \sup_{\pol \in \Pols} \inf_{\bel \in \Bels} U(\bel, \pol)
  =
  \inf_{\bel \in \Bels}  \sup_{\pol \in \Pols}  U(\bel, \pol)
  = U_*.
\end{equation}
Let $\pol^*$ and $\bel^*$ be the maximin policy and minimax prior respectively.
If the game has a value then there exists an equalising policy which is optimal for some belief $\bel^*$, and vice versa. A sufficient condition for this to occur is for $U^*$ to be convex and differentiable everywhere~\citep[c.f.][]{grunwald-dawid:game-robust-bayesian:aos:2004}. In order to study when this can occur, we first go over a couple of well-known facts.

\section{Existence of maximin policies}

\begin{definition}[Policy]
  Let $\CH$ be the set of all histories $(s_t,\, a_{t-1})$. A (stochastic) policy  $\pol$ is a set of probability measures $\cset{\pol(\cdot \mid h)}{h \in \CH}$ on the set of actions $\CA$. We denote the set of all (history-dependent, stochastic) policies by $\PolsHS$.
\end{definition}
\begin{definition}[Deterministic policy]
  A policy is deterministic if, for each sequence $s^t,\, a^{t-1}$, there exists an action $a \in \CA$ such that $\pol(a_t = a \mid s_t,\, a_{t-1}) = 1$. We denote the set of deterministic policies by $\PolsHD$.
\end{definition}
\begin{definition}[Memoryless policy]
  A policy is memoryless (or reactive) if, for all sequences $s_t, a_{t-1}$, we have $\pol(a_t = a\mid s_t,\, a_{t-1}) = \pol(a_t = a \mid s_t)$. We denote the set of  memoryless (stochastic) policies by $\PolsMS$.
\end{definition}
The set of memoryless deterministic policies is denoted by $\PolsMD$. 
Obviously, $\PolsMD \subset \PolsHD \subset \PolsHS$ and $\PolsMD \subset \PolsMS \subset \PolsHS$.

\begin{definition}[Mixed policy]
  A mixed policy is a probability measure over policies. If $\Pols$ is a set of base policies, we denote the set of probability measures over $\Pols$ by $\Simplex(\Pols)$.
\end{definition}
\begin{fact}
  For any MDP $\mdp$ there exists a deterministic, memoryless policy that is optimal, i.e.
$U^*(\mdp) = \sup_{\pol \in \Pols} U(\mdp, \pol) = \max_{\pol \in \PolsMD} U(\bel, \pol)$.
\end{fact}
\begin{fact}
  For any distribution $\bel$ over MDPs, there exists a deterministic, history-dependent policy that is optimal, i.e.
$U^*(\bel) = \sup_{\pol \in \Pols} U(\bel, \pol) = \max_{\pol \in \PolsHD} U(\bel, \pol)$
\end{fact}
A well-known game theoretic result is that an equalising policy can always be found in $\Simplex(\Pols)$ when $\Pols$ is finite. However for fixed, finite $T$, there may not exist a deterministic equalising policy. Then the number of possible policies is finite and consequently $U^*$ is piecewise-linear.  
\begin{remark}
  If $\PolsHD$ is finite, there exists a policy $\pol^* \in \PolsHS$ achieving the value of the game.
\label{thm:phs-value}
\end{remark}
\begin{proof}
  For any mixed policy $\delta \in \Simplex(\PolsHD)$, there exists an equivalent stochastic policy in $\PolsHS$. This can be constructed by augmenting the state space to include the outcomes of fair coins.
  Now note that there exists an optimal mixed strategy $\delta^* \in \Simplex(\PolsHD)$ achieving the value of the game, as the number of finite-horizon policies is finite. But there also exists a stochastic policy with the same distribution for all histories $h \in \CH$. 
\end{proof}

\citet{grunwald-dawid:game-robust-bayesian:aos:2004} make some interesting connections between maximum entropy and robust Bayesian decisions. In particular, they define the generalised entropy of a distribution $\bel$ to be minimum loss $H(\bel) \defn \inf_\pol L(P, \pol)$ achievable in a game between nature and a decision maker. In our setting, it is natural to consider the following two loss functions:
\begin{equation}
  \label{eq:loss-function}
  L_1(\bel, \pol) = \beta^\top (V^*_\bel - V^\pol_\bel),
  \qquad
  L_2(\bel, \pol) = \beta^\top \E_\bel (V^*_\mdp - V^\pol_\mdp),
\end{equation}
where $\beta \in \Reals^{|\CS|}$ is a distribution on the states. These corresponds to the regret of $\pol$ relative to the $\bel$-optimal policy and to the oracle policy respectively. Now, let $\Bels$ be a set of probability distributions on $\MDPs$.  One idea is to try and guard against the worst-case  prior in a restricted set $\Bels_\hstat$, by constraining the expectation of a statistic $\stat : \MDPs \to \Reals^k$ under the prior to be equal to the observed value of the statistic, $\hstat$.
\begin{equation}
  \label{eq:restricted-prior}
  \Bels_{\hstat} = \cset{\bel \in \Bels}{\Eb(\stat) = \hstat}.
\end{equation}
One possibiltiy is to use the cumulative state distribution for a particular policy $\pol$, i.e.
\begin{equation}
  \stat(\mdp) = (I - \gamma P_\mdp^{\pol})^{-1},
  \label{eq:statistic}
\end{equation}
where a common choice for $\pol$ is the optimal policy for MDP $\mdp$, $\pol^*(\mdp)$, used for example in~\citet{mannor:empirical-bayes-envelope-cMDP}.  In that case it is easy to see that $V_\bel^* = \Eb(\stat) \rew$ . A sufficient condition for a policy $\pol$ to be robust Bayes against $\Bels_{\hstat}$ is for it to be linear, that is
$L(\mdp, \pol) = \alpha_0 + \alpha^\top \stat(\mdp)$ for all $\mdp \in \MDPs$. Then it is also true~\citep[see][Theorem 7.1]{grunwald-dawid:game-robust-bayesian:aos:2004} that $\pol$ is an equalising policy against $\Bels_\hstat$. 

How can $\Bels_\hstat$ be calculated? When the set of MDPs is finite, then it is defined through the linear equation $\Bels_\hstat = \cset{\bel}{\sum_\mdp \bel(\mdp) \stat(\mdp) = \hstat}$. Given, then, a sequence of observations, a statistic, and a resulting set of priors $\Bels_{\hstat}$, one important question is how we can efficiently calculate such policies. We explain this in the following section.

\section{Calculating robust policies}

One potential solution involves finding the minimax prior $\bel^*$ and then the policy that is Bayes-optimal with respect to it. In previous work~\citet{Koolen:MSc} has shown that finding $\bel^*$ can be found via a concave-linear optimisation, for the `truth-finding' game. However, this is not generally true. However, it has been shown by \citet{freund1999adaptive}, that  multiplicative weighs algorithm can be used for zero-sum games, as long as an oracle that can compute best responses is available. In our setting, this would correspond to nature having the ability to efficiently construct a worst-case MDP given a policy. 
On the other hand~\citet{fictitious-stochastic-games} show that the value in stochastic zero sum games can be achieved asymptotically via fictitious play even when the game matrix is known approximately, as long as the approximation converges to the true game matrix.

One idea is to apply results from the experts literature, such as the weighted majority algorithm (WMA). We start by assuming that in each round $k$ the Decision Maker has full access to the information regarding the rewards of the past round. That means that she can observe the outcomes of \emph{all} the policies that were available previously. 

Assume that $|\CM| = M$ is finite\footnote{If $M=\infty$, a grid over the MDPs can be used to obtain a finite set. Then the approximation error for an MDP that has an $\epsilon$-close transition matrix and mean reward from an MDP on the grid, will be bounded by $\epsilon/(1-\gamma)^2$.}. Then and let $\bv_\pol$ be the $1\times M$ vector of values for policy $\pol$ for the given set of Markov decision problems $\MDPs$:
\[
\bu^\pol=\left(U(\mu, \pol)\right)_{\mdp \in \MDPs}.
\]
Assume that $|\Pols| = N$ is finite and denote by $\bu_{\xi}$ the $1\times N$ vector of values for each policy
\[
	\bu_\xi=\left(U(\xi, \pol)\right)_{\pol \in \Pols}.
\]
To apply WMA (Alg.~\ref{alg:wma}) we execute policies in rounds. At each round, we select a policy $\pol_{(k)}$ from a distribution $\Qk$, and nature calculates chooses some prior $\bel_{(k)}$. 
Moreover, denote by $x_{\pol_i,k}$ the \emph{total realized reward} obtained by following policy $\pol_i$, $i=1,2,...,N$, in the $k$-th round. Each $x_{\pol_i,k}$ is a random variable that has an expected value, equal to $U(\xi,\pol_i)$.
Finally, denote by $\textbf{x}_{(k)}$ the vector of sampled utilities of all policies up to round $k$:
\[
\textbf{x}_{(k)} = (x_{\pol_1,k},\ x_{\pol_2,k},\ \ \dots \ ,\ x_{\pol_N,k}).
\]

\begin{algorithm}[H]
\textbf{Input}: 
	 A set of policies $\pol$, with ${|\pol|}=N$; a set of weights $w_{(k)}=\big(w_{i,k}\big)_{i=1}^N$; a learning rate $0<\ell\leq1/2$.
\textbf{Initialize}: $w_{i, 1}=1$.
\caption{WMA}
\label{alg:wma}
\textbf{For} each round $k$:

	\begin{algorithmic}[1]

		\State DM(Decision Maker) normalizes the weights to get a distribution $\Qk=\frac{w_{(k)}}{\sum_{i=1}^N w_{i,k}}$

		\State  DM selects $\pol_{(k)}$  among  $\pol_i, \ i=1,2,...,N$ according to the distribution $\Qk$

		\State Nature chooses $\displaystyle \xi_{(k)}\in  \text{argmin}_{\xi_{(k)}}\mathbbm{E}_{\Qk}\left[ \bu_{\xi_{(k)}} \right ]$

		\State DM receives reward $x_{k,\pol_{(k)}}$ and calculates  $U(\xi_{(k)},\pol_i)$ for all policies $\pol_i\in\pol$

		\State DM calculates the next set of weights for $i=1,\ \dots \ ,\ N$:
		\[w_{i, k+1}=\left(1+\ell \,U(\xi_{(k)}, \pol_i) \right)w_{i,(k)}\]
		
	\end{algorithmic}
\end{algorithm}

\subsection{Analysis}

	The main issue is the computation of the value $U(\xi_{(k)}, \pol_i)$, which is used in steps 3 and 4. Depending on the sizes of the policy and MDP space accurate or approximate values for the quantity $U(\xi_{(k)},\pol_i)$ can be obtained.

\paragraph{When $N$ (number of policies) and $M$ (number of MDPs) are small.} Then we can retrieve the expected value of each policy $V_\mu^{\pol_i}=\left(I-\gamma P_\mu^{\pol_i}\right)^{-1} \rew$, where $P_\mu^{\pol_i}$ is the kernel. The inverse operator will require $O(|\CS|^3)$.
	The expected reward for sampling a policy $\pol$ from the distribution $\Qk$ is
	\[
	\mathbbm{E}_{\pol\sim\Qk}\left[x_{k,\pol}\right] = \textbf{x}_{(k)}\cdot \Qk.
	\]
	The total expected reward over all rounds is therefore 
	\[
	\mc{V}_{\text{WMA}}^{(K)}\triangleq \sum_{k=1}^K \textbf{x}_{(k)}\cdot \Qk.
	\]
		\begin{theorem}[\cite{arora2012multiplicative}]
The Multiplicative Weights algorithm guarantees that after $K$ rounds, for any distribution $\mc{Q}$ on the decisions, it holds:
	\[
	\mc{V}_{\text{WMA}}^{(K)} \geq \sum_{k=1}^K\left(\bu_{\xi_k}-\ell\vert \bu_{\xi_k}\vert\right)\cdot\mc{Q}-\frac{\log_eN}{\ell}
	\]
	where $\vert \bu_{\xi_k}\vert$ is the vector obtained by taking the coordinate-wise absolute value of the vector containing the expected values $V_{\xi_k}^{\pol_i}$.
 	\end{theorem}

\paragraph{If $M$ is small, but $N$ is large} then computation of $U(\xi_{(k)}, \pol_i)$ becomes difficult. We can approximate the true value of the expected values, though, by using a Monte Carlo sampling of $S$ iterations. We substitute  $U(\xi_{(k)}, \pol_i)$ with its estimator $\hat U(\xi_{(k)}, \pol_i, S)$ (lines 3,4 \& 5 of   algorithm WMA) and we call this modification of the algorithm as WMA-SR. Moreover, for each round $k$, we need to introduce an estimation error term  $\mc{E}_{(k)}^{(S)}$ (which depends on the number $S$ of Monte Carlo iterations), since now the weights are updated by using approximations and not the true values.

\begin{algorithm}[H]
\textbf{Input}: 
	 A set of policies $\pol$, with ${|\pol|}=N$; a set of weights $w_{(k)}=\big(w_{i,k}\big)_{i=1}^N$; a learning rate $0<\ell\leq1/2$.
\textbf{Initialize}: $w_{i, 1}=1$.
\caption{WMA-SR}
\label{alg:wma}
\textbf{For} each round $k$:

	\begin{algorithmic}[1]

		\State DM(Decision Maker) normalizes the weights to get a distribution $\Qk=\frac{w_{(k)}}{\sum_{i=1}^N w_{i,k}}$

		\State  DM selects $\pol_{(k)}$  among  $\pol_i, \ i=1,2,...,N$ according to the distribution $\Qk$

		\State Nature chooses $\displaystyle \xi_{(k)}\in  \text{argmin}_{\xi_{(k)}}\mathbbm{E}_{\Qk}\left[ \bu_{\xi_{(k)}} \right ]$

		\State DM receives reward $x_{k,\pol_{(k)}}$ and calculates  $\hat U(\xi_{(k)}, \pol_i, S)$ for all policies $\pol_i\in\pol$

		\State DM calculates the next set of weights for $i=1,\ \dots \ ,\ N$:
		\[w_{i, k+1}=\left(1+\ell \,\hat U(\xi_{(k)}, \pol_i, S) \right)w_{i,(k)}\]
		
	\end{algorithmic}
\end{algorithm}

		The  value earned by using WMA-SR, over all rounds is 
	\[
	\mc{V}_{\text{WMA-SR}}^{(K)}\triangleq\E\left(\sum_{k=1}^K \textbf{x}_{(k)}\cdot \Qk\right)=\sum_{k=1}^K \bu_{\xi_{(k)}}\cdot\Qk
	\]
	where $\Qk=\left(q_{k,1},\dots,q_{k,N}\right)$.
	
	 First, we prove a lemma for the approximated expected values.
	
		\begin{lemma}\label{thm:1}
	Assume that all policy rewards lie in $[-1,1]$. Let $0<\ell\leq\frac{1}{2}$. Then  after $K$ rounds, it holds:
	\[
	\mc{V}_{\text{WMA-SR}}^{(K)}\geq \sum_{k=1}^{ K}\hat U(\xi_{(k)}, \pol_i, S)-\ell\sum_{k=1}^K\vert \hat U(\xi_{(k)}, \pol_i, S)\vert-\frac{\log_eN}{\ell}
	\]
	for all $i=1,2,...,N$.
 	\end{lemma}
Observe that if the rewards are not stochastic, then Lemma \ref{thm:1} is reduced to the standard expert setting\citep[c.f]{arora2012multiplicative,freund1999adaptive}.
\begin{theorem}\label{thm:wma1} 
 	Assume that all policy rewards  lie in $[-1,1]$. Let $0<\ell\leq\frac{1}{2}$. Let $\varepsilon>0.$ Then  after $K$ rounds, for the total expected rewards, it holds:
	\[
	\mc{V}_{\text{WMA-SR}}^{(K)} \geq \sum_{k=1}^{K} U(\xi_{(k)}, \pol_i)-\ell\sum_{k=1}^K\vert  U(\xi_{(k)}, \pol_i)\vert-\frac{\log_eN}{\ell}-\sum_{k=1}^K\mc{E}_{(k)}^{(S)}
	\]
	for all $i=1,2,...,N$, where $\xi_{(k)}=(\xi_{1,k},\dots,\xi_{M,k})$, $\sum_m \xi_{m,k}=1$, $\Qk=(q_{1,k},\dots,q_{N,k}), \sum_i q_{i,k}=1$, $\mc{E}_{(k)}^{(S)}$ is the error term of the $k$-th round (and $S$ denotes the number of Monte Carlo simulations): 
\[
	\mc{E}_{(k)}^{(S)}= \Big\vert   \bu_{\xi_{(k)}}\cdot\Qk - U(\xi_{(k)}, \pol_i)+\ell\vert  U(\xi_{(k)}, \pol_i)\vert+\frac{\log_eN}{\ell}
	-\left( \hat\bu_{\xi_{(k)}}\cdot\Qk - \hat U(\xi_{(k)}, \pol_i, S)+\ell\vert \hat U(\xi_{(k)}, \pol_i, S)\vert+\frac{\log_eN}{\ell}\right) \Big\vert
\]
and
\[ \Pr\left(\left\vert\sum_{k=1}^K\mc{E}_{(k)}^{(S)}<\varepsilon\right\vert\right) \geq 1-2\exp\left(-k\frac{\varepsilon^2}{2}\right).
\]

	\end{theorem}
The error bound is retrieved by applying Azuma's Lemma, since $\hat U(\xi_{(k)}, \pol_i, S)- U(\xi_{(k)}, \pol_i)$ is a martingale difference and all rewards lie in $[-1,1]$.
We can also obtain a result for a distribution $\mc{P}$ over $\pol_i$'s, $i=1,2,...,N$.
\begin{corollary} \label{cor:wma11}
After $K$ rounds, for any distribution $\mc{P}\in\mathbbm{R}^{N\times 1}$ on the decisions, it holds:
	\[
	\sum_{k=1}^K \bu_{\xi_{(k)}}\cdot\Qk \geq \sum_{k=1}^K\left(\bu_{\xi_{(k)}}-\ell\vert \bu_{\xi_{(k)}}\vert\right)\cdot\mc{P}-\frac{\log_eN}{\ell}-\sum_{k=1}^K\mc{E}_{(k)}^{(S)}
	\]
	where $\vert V_{\xi_{(k)}}\vert$ is the vector obtained by taking the coordinate-wise absolute value of $\bu_{\xi_{(k)}}$.
 	\end{corollary}
	
\begin{definition}
The regret of the learning algorithm against the optimal distribution $\mc{P}^\star\in \text{argmax}_P\{\bu_{\xi_{(k)}}\cdot\mc{P}\} $ is 
\[
B(K)=\sum_{k=1}^K \bu_{\xi_{(k)}}\cdot\mc{P}^\star - \sum_{k=1}^K \bu_{\xi_{(k)}}\cdot\Qk
\]\end{definition}

Corollary \ref{cor:wma11} can be used in order to bound the regret. To that end, we first need the following theorem.

\begin{theorem}\label{thm:iznogud}
After $K$ rounds of applying the modified weighted majority algorithm WMA-SR, for any distribution $\mc{P}$ it holds:
\[
\sum_{k=1}^K \bu_{\xi_{(k)}}\cdot\mc{P} - \sum_{k=1}^K \bu_{\xi_{(k)}}\cdot\Qk\leq2\sqrt{\log_e N K}+\sum_{k=1}^K\mc{E}_{(k)}^{(S)}
\]
\end{theorem}

\begin{corollary} \label{cor:iz}
When algorithm WMA-SR is run with parameter $\ell=\sqrt{\frac{\log_e N}{K}}$ then the regret of the algorithm is bound by
$B(K)\leq2\sqrt{\log_e N K}+\sum_{k=1}^K\mc{E}_{(k)}^{(S)}$.
\end{corollary}
One can also show that the algorithm converges  by dividing the time into epochs. A choice of epochs that gives convergence with probability one is to define the length of each epoch as $T_k=k^2$
and then proceed similarly to \cite{freund1999adaptive} to show the following
\begin{theorem} Suppose we repeat the game for an unbounded number of rounds. Then for the regret of the algorithm it holds: $\Pr\left[\frac{B(K)}{K}\leq \epsilon\right]=1$
for all but a finite number of values of $K$ and for every $\epsilon>0$.
\end{theorem}
\paragraph{View as a bandit problem.} To avoid uniformly sampling all policies, we could cast this as a contextual bandit problem, mapping the prior $\bel_k$ selected at each round by nature to the context. With a slight modification of the linear context bandit algorithm presented in~\cite{Auer2002using}, we then obtain a similar bound, which is however linear in the number of policies, making this approach impractical. Since our estimates converge to the values of the policies, we can recover the value of the game if Nature uses fictitious play with respect to our estimates~\citep{fictitious-stochastic-games}.

\section{Conclusion}

We have discussed the links between robust reinforcement learning and maximin policies. In particular, an interesting idea to explore is to use maximin policies against a constrained set of priors $\Bels_{\hstat}$. Such as set is easy to define for a finite number of MDPs. However, to put the idea into practice we also need to establish computational procedures for approximately calculating such policies. Although this seems to be achievable for small problems, there does not appear to be a useful procedure for larger ones. One potential direction would be to choose statistics $\stat$ that inherently make the problem amenable to simple solutions.

\bibliographystyle{plainnat}
\bibliography{references}

\end{document}